\documentclass[runningheads]{llncs}
\usepackage[T1]{fontenc}
\usepackage{graphicx}
\usepackage{subcaption}  
\begin{document}
\title{Leveraging Graph Retrieval-Augmented Generation to Support Learners' Understanding of Knowledge Concepts in MOOCs}
\titlerunning{Leveraging Graph RAG to Support Learners}
%

\author{Mohamed Abdelmagied \and
Mohamed Amine Chatti \and
Shoeb Joarder \and
Qurat Ul Ain \and
Rawaa Alatrash}
\authorrunning{M. Abdelmagied et al.}
\institute{Social Computing Group, Faculty of Computer Science, University of Duisburg-Essen,  Germany\\
\url{https://www.uni-due.de/soco/}}
\maketitle              
\begin{abstract}

Massive Open Online Courses (MOOCs) lack direct interaction between learners and instructors, making it challenging for learners to understand new knowledge concepts. Recently, learners have increasingly used Large Language Models (LLMs) to support them in acquiring new knowledge. However, LLMs are prone to hallucinations which limits their reliability. Retrieval-Augmented Generation (RAG) addresses this issue by retrieving relevant documents before generating a response. However, the application of RAG across different MOOCs is limited by unstructured learning material. Furthermore, current RAG systems do not actively guide learners toward their learning needs. To address these challenges, we propose a Graph RAG pipeline that leverages Educational Knowledge Graphs (EduKGs) and Personal Knowledge Graphs (PKGs) to guide learners to understand knowledge concepts in the MOOC platform CourseMapper. Specifically, we implement (1) a PKG-based Question Generation method to recommend personalized questions for learners in context, and (2) an EduKG-based Question Answering method
that leverages the relationships between knowledge concepts in the EduKG to answer learner selected questions. To evaluate both methods, we conducted a study with 3 expert instructors on 3 different MOOCs in the MOOC platform CourseMapper. The results of the evaluation show the potential of Graph RAG to empower learners to understand new knowledge concepts in a personalized learning experience.  

\end{abstract}
\section{Introduction}
Massive Open Online Course (MOOC) platforms have emerged as a key digital technology driving the change in the educational landscape in the last decade, as they promote self-regulated and lifelong learning \cite{haleem2022understanding}. However, MOOC platforms present learners with new challenges due to lack of interaction with their instructors. Therefore, learners might find it difficult to acquire new knowledge that helps them achieve their goals \cite{henderikx2021making}.
Recently, Large Language Models (LLMs) have shown remarkable performances in numerous Natural Language Processing (NLP) tasks such as Question Answering \cite{naveed2023comprehensive}. Their ability to answer general knowledge questions led to extensive use by learners to answer their questions in education. However, there is concern about the unsystematic use of these models, as it has several practical issues, such as hallucinations \cite{wang2024large}. Hallucinations lead LLMs to generate context-unaware or even false content, which can have negative effects on learners. Therefore, several educational systems have emerged that try to mitigate these issues using methodologies such as Retrieval-Augmented Generation (RAG). 
RAG aims to reduce hallucinations by retrieving relevant documents from a knowledge base and providing them to the LLM before generating a response \cite{dan2023educhat,liu2024teaching,liu2024hita}. 

However, the complex structure and relationships between different concepts in knowledge bases can represent a challenge for RAG systems. Graph RAG \cite{peng2024graphretrievalaugmentedgenerationsurvey} can address this challenge by its ability to capture complex relationships between different concepts and establishing links between different documents in Knowledge Graphs (KGs) to retrieve more relevant and personalized information.
Graph RAG can be divided into three main steps, which are \textit{graph-based indexing}, \textit{graph-guided retrieval} and \textit{graph-enhanced generation} \cite{peng2024graphretrievalaugmentedgenerationsurvey}. \textit{Graph-based indexing} stores KG data in a manner that allows efficient traversal of graph information, which enhances retrieval quality. \textit{Graph-guided retrieval} leverages structural information in a KG to retrieve more accurate results. \textit{Graph-enhanced generation} uses the information provided from the KG to generate a response with better reasoning.

Another limitation of RAG that causes its application to be limited to specific courses is that it requires the creation of knowledge bases from course materials that are specifically curated for retrieval. However, this is not always the case in MOOCs, where instructors often upload unstructured materials such as lecture slides that may lack the necessary structure for effective knowledge retrieval. Moreover, this constraint limits learners from accessing additional resources that could provide deeper insights beyond the content presented in the MOOC.
Educational Knowledge Graphs (EduKGs) can be an effective tool to extract structured knowledge concepts from learning material and linking to external learning resources that can be used as a supporting data source in RAG \cite{abu2025llm}. 

RAG systems in education have an additional limitation in that they require learners to pro-actively formulate questions according to their learning needs \cite{liu2024hita,liu2024teaching,dan2023educhat}. However, they do not actively guide learners in identifying the specific knowledge or questions required to achieve their learning goals and do not direct them in learning the essential knowledge concepts necessary for their educational progress. This can cause learners to diverge from their learning goals or ask off-topic questions that would cause the system to hallucinate. Therefore, it is essential for such systems to guide learners by indicating what knowledge concepts they need to learn and providing personalized recommendations of questions that would help them understand the knowledge concepts. 
Asking the right questions requires an effective modeling of the learners' needs. Recently, Personal Knowledge Graphs (PKGs) have been proposed to model learners in a MOOC context, based on the knowledge concepts that they did not understand \cite{ain2024learner}. These PKGs can help learners identify knowledge concepts they need to learn to achieve their goals. Moreover, PKGs can be leveraged to generate personalized questions that can help learners understand new knowledge concepts.

In this paper, we leverage EduKGs and PKGs to implement a Graph RAG pipeline that can guide learners in understanding new knowledge concepts in the MOOC platform CourseMapper \cite{ain2022learning}. 
Specifically, we propose (1) a PKG-based Question Generation method to recommend personalized questions for each learner according to the knowledge concepts that they do not
understand, and (2) an EduKG-based Question Answering method to answer user-selected questions by leveraging the relationships between knowledge concepts in the EduKG. Furthermore, we evaluated the two proposed methods in our Graph RAG pipeline with three expert instructors on three different MOOCs to assess the linguistic and task-oriented qualities of the generated questions and answers. Our study demonstrates the potential of our proposed Graph RAG pipeline to empower learners to control the Question Generation and Answering processes in MOOCs, according to their learning needs and goals. In particular, leveraging PKGs to guide learners to ask questions that help them understand the knowledge concepts was perceived as effective to help learners ask the right questions in context. 
\section{EduKG and PKG Construction in CourseMapper}\label{kg_construction}
\begin{figure}[h!]
    \centering
    \includegraphics[scale=0.4]{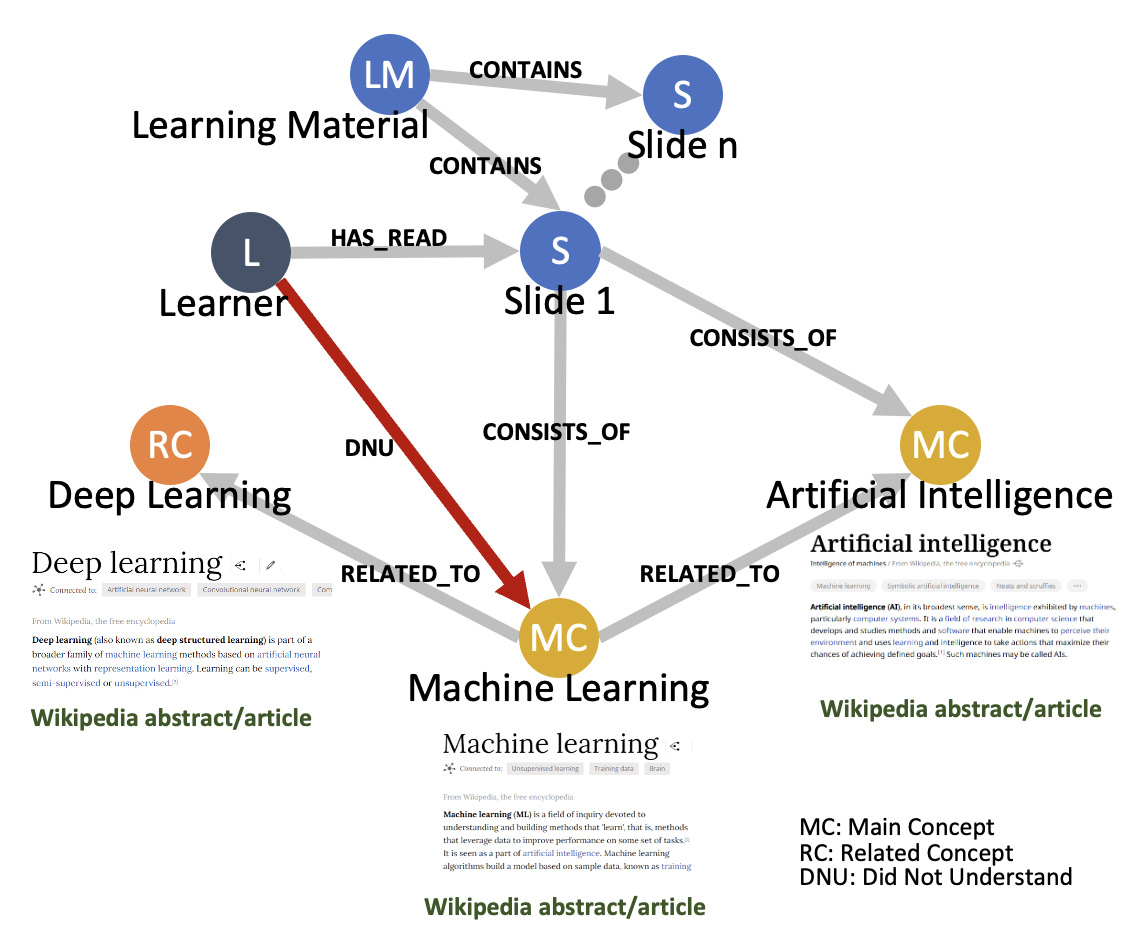} 
    \caption{An overview of an example EduKG in CourseMapper: Each Learning Material (LM) contains Slides (S), Each Slide consists of Main Concepts (MCs) which also correspond to Wikipedia Articles. Each MC is related to Related Concepts (RCs) which are further concepts extracted from the MC article on Wikipedia.}
    \label{EduKG_MOOC}
\end{figure}

In general, a KG is a graph in which the nodes are knowledge entities and the edges are relationships between them. An EduKG can represent any type of entities in educational systems. These entities can be knowledge concepts, instructors, courses, or even universities \cite{ain2023automatic}. In CourseMapper, instructors can upload Learning Materials (LM) to the MOOC platform. For every LM, an EduKG is automatically constructed, where the EduKG includes further entities such as Slides (S), Main Concepts (MC), Related Concepts (RC) and Learners (L), as shown in Figure \ref{EduKG_MOOC}. Each LM contains a set of S. Then, each S consists of a set of MCs extracted using a keyphrase extraction algorithm. These MCs are tagged with Wikipedia articles that discuss the concept. The EduKG is expanded by extracting other Wikipedia articles referenced in the MC and adding them as RCs. Furthermore, learners can update their knowledge states by identifying MCs as "Did Not Understand" (DNU). By allowing learners to update their states, learners can personalize the EduKG to produce their PKG. Both the EduKG and PKG information are stored in a Neo4j graph database. The EduKG and PKG give learners a structured overview of what concepts should be learned to achieve their learning goals. 
\section{Methodology} \label{methodology}
In this section, we present our approach to achieve personalized Question Generation and Question Answering using Graph RAG to support learners' understanding of new knowledge concepts in CourseMapper. Firstly, we present the abstract pipeline by referring to a user scenario. Then, we dive into the technical aspects of the pipeline by discussing our workflows for implementing PKG-based Question Generation and EduKG-based Question Answering.
\subsection{User Scenario} 
\begin{figure}[h!]
    \centering
    \scalebox{0.55}{
        \begin{minipage}{\textwidth}
            \centering
            \begin{subfigure}{0.49\textwidth}
                \centering
                \includegraphics[width=\textwidth]{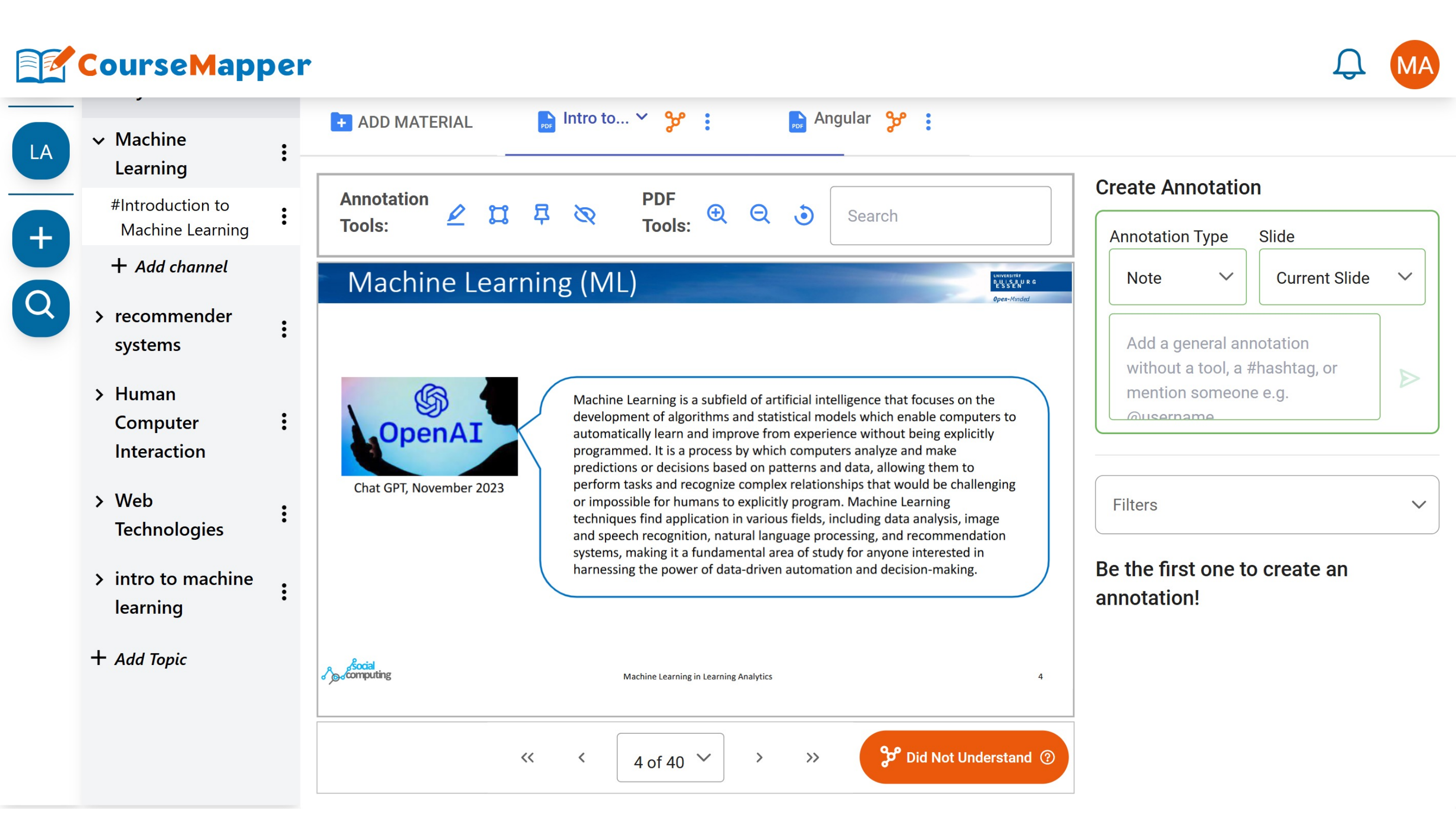}
                \caption{Select 'Did Not Understand'}
            \end{subfigure}
            \hfill
            \begin{subfigure}{0.49\textwidth}
                \centering
                \includegraphics[width=\textwidth]{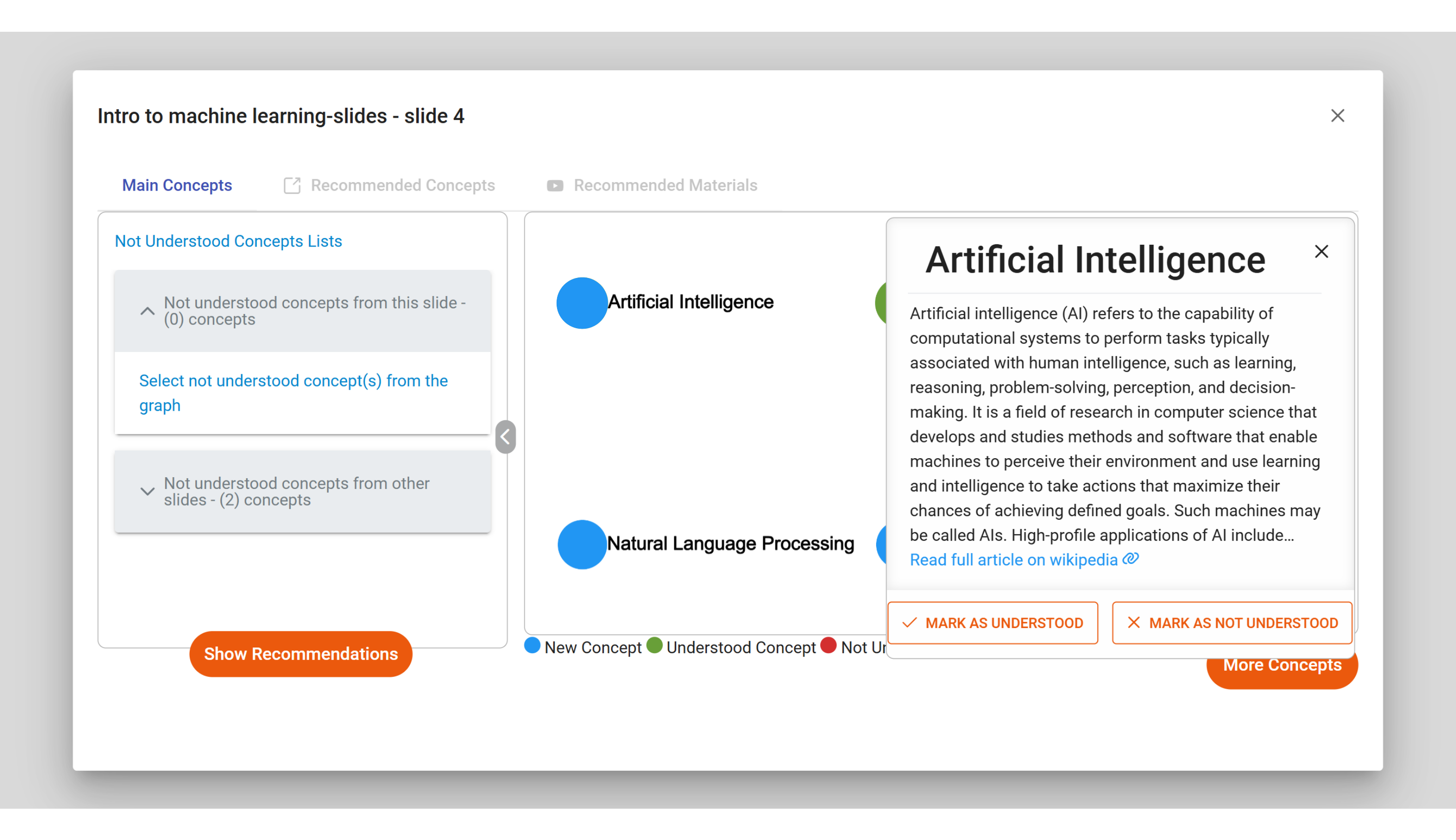}
                \caption{Mark 'Artificial Intelligence' as DNU}
            \end{subfigure}

            \vspace{1em} 

            \begin{subfigure}{0.49\textwidth}
                \centering
                \includegraphics[width=\textwidth]{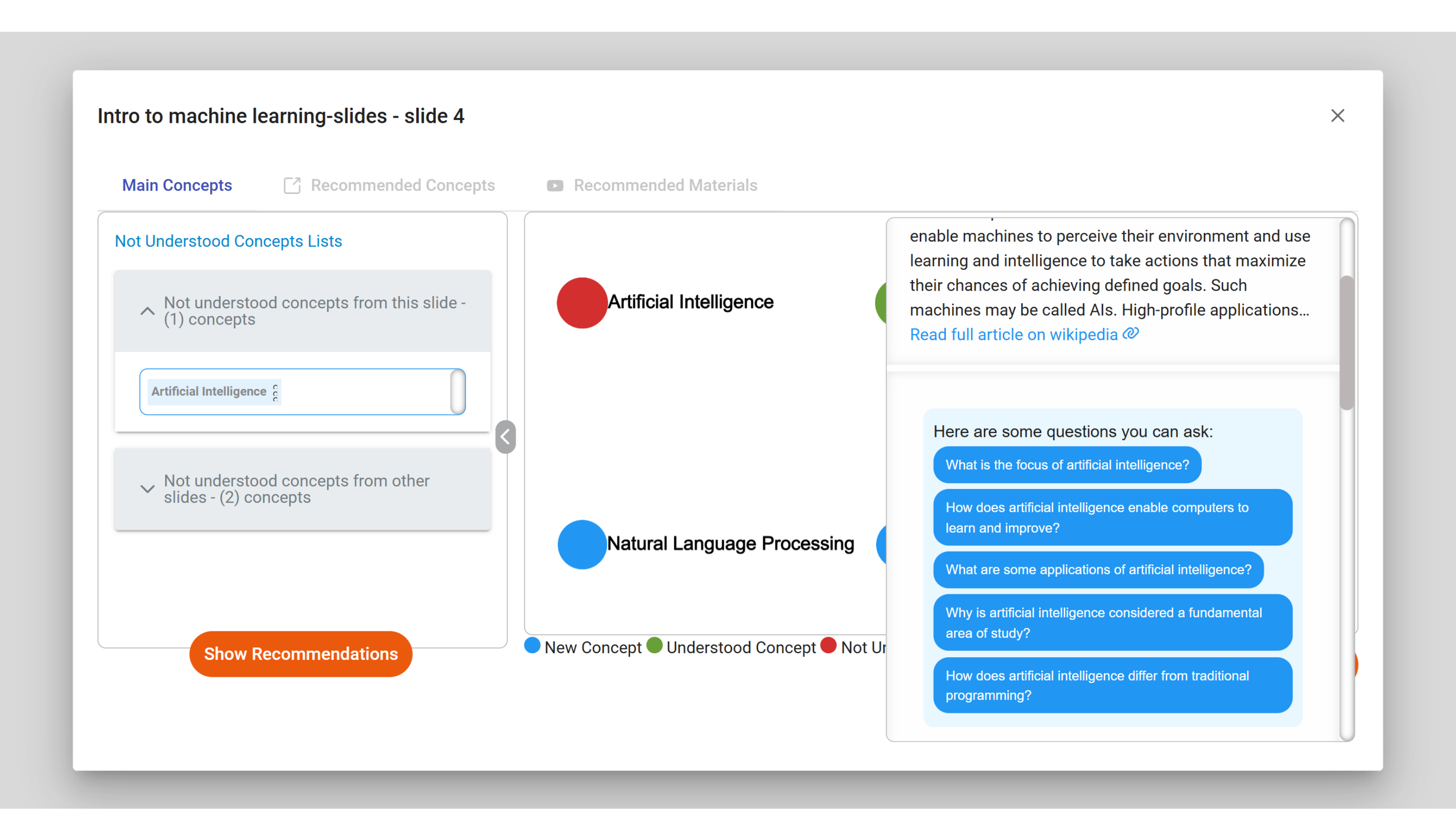}
                \caption{Question Generation for DNU Concept}
            \end{subfigure}
            \hfill
            \begin{subfigure}{0.49\textwidth}
                \centering
                \includegraphics[width=\textwidth]{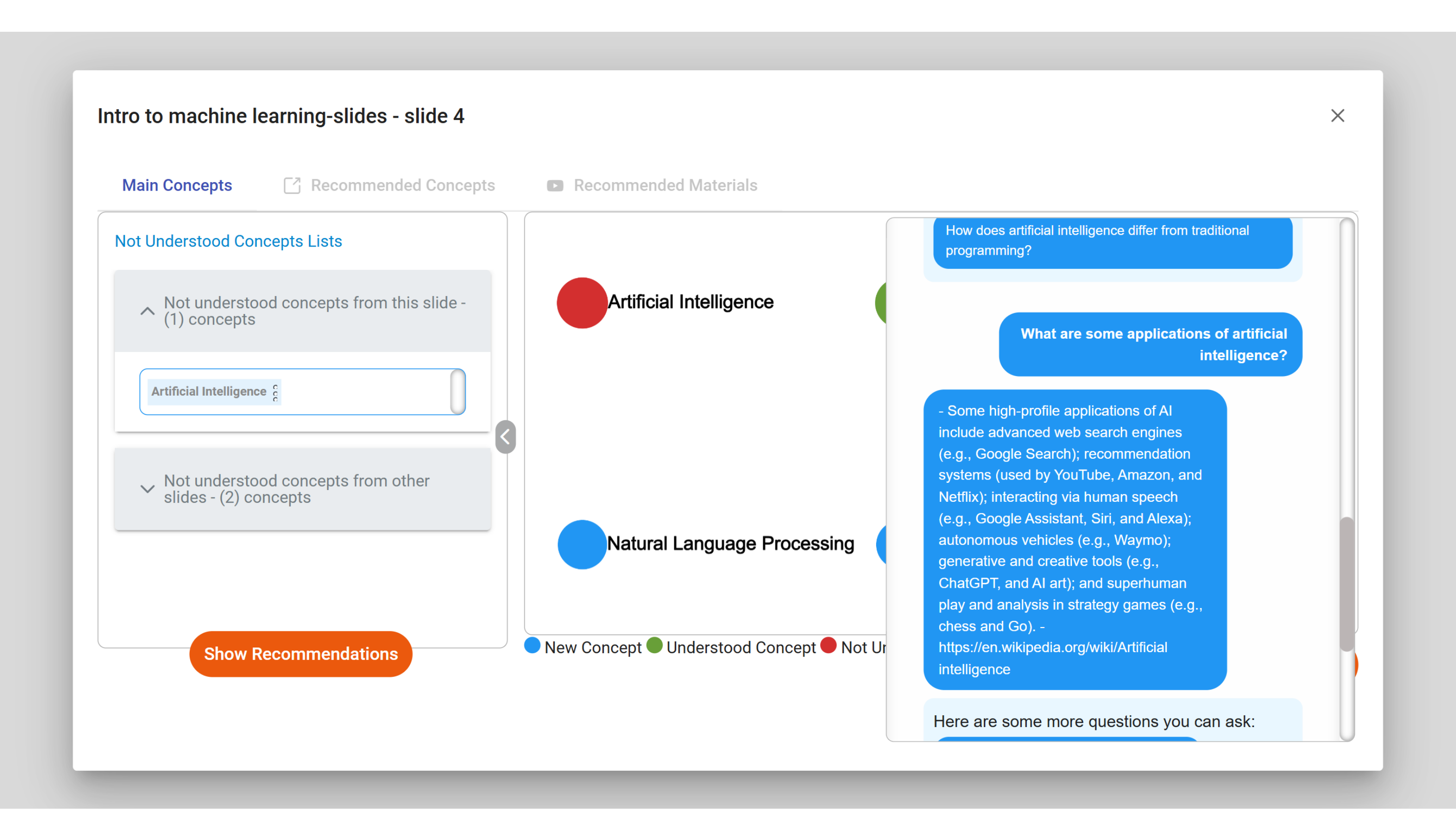}
                \caption{Question Answering and Citation}
            \end{subfigure}
        \end{minipage}
    }

    \caption{A user scenario of the PKG-based Question Generation and EduKG-based Question Answering in CourseMapper}
    \label{user-scenario-fig}
\end{figure}

Farah is a university student who is enrolled as a learner in the MOOC 'Learning Analytics' delivered through CourseMapper. The instructor of the MOOC uploaded a new LM titled 'Introduction to Machine Learning'. While navigating the slides in the LM, she recognizes that she is unable to understand 'Slide 4', which mentions the definition of Machine Learning (Figure \ref{user-scenario-fig},\textbf{a}). Therefore, she views her PKG for that slide. She notices several MCs for 'Slide 4' (Figure \ref{user-scenario-fig},\textbf{b}). She realizes that she still does not fully understand the concept of 'Artificial Intelligence', which is a broader field than Machine Learning. She selects the concept and clicks on 'MARK AS NOT UNDERSTOOD'. Afterward, she sees a dialog open with some suggested questions that can help her understand the concept of 'Artificial Intelligence' as shown in Figure \ref{user-scenario-fig} \textbf{c}. Therefore, she selects the third question 'What are some applications of artificial intelligence'. Finally, she receives answers \ref{user-scenario-fig} \textbf{d}, which are supported by evidence from a Wikipedia article in the EduKG. Now, Farah understands more about 'Artificial Intelligence' and is motivated to explore more about this concept by selecting other generated questions.

\subsection{PKG-based Question Generation} \begin{figure}[h!]
    \centering
    \includegraphics[width=\textwidth]{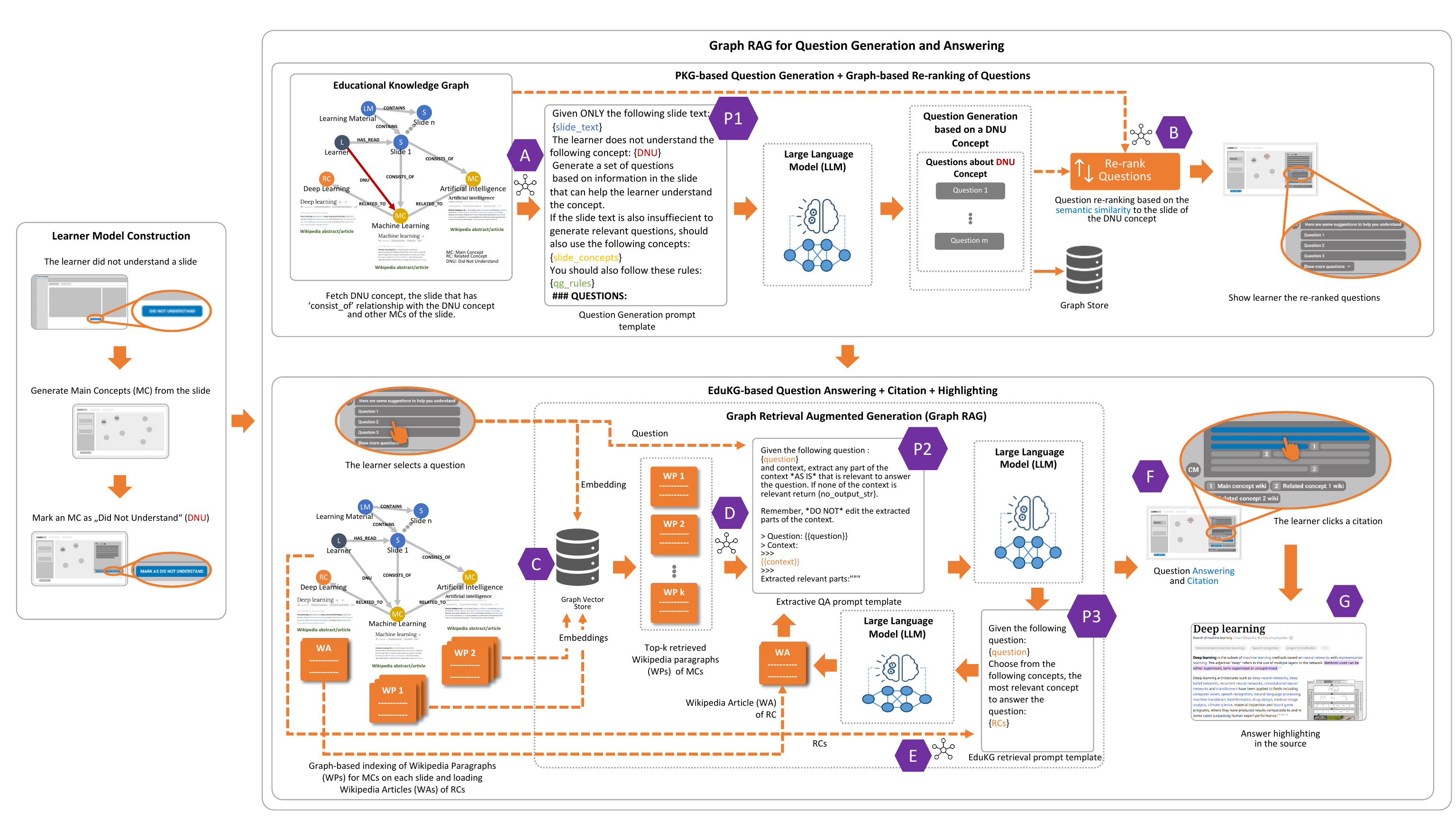} 
    \caption{An overview of the pipeline for implementing PKG-based Question Generation and EduKG-based Question Answering with the steps: (\textbf{A}) \textit{graph-guided retrieval} for Question Generation, (\textbf{P1}) Question Generation prompt template, (\textbf{B}) \textit{graph-based re-ranking},  (\textbf{C}) \textit{graph-based indexing}, (\textbf{D}) and (\textbf{E}) \textit{graph-guided retrieval} for Question Answering, (\textbf{P2}) Extractive Question Answering prompt template, (\textbf{P3}) EduKG retrieval prompt template, (\textbf{F})
    Answers with citations, (\textbf{G}) Highlighted answers}
    \label{pipeline-1-fig}
\end{figure}   
To generate questions that address individual learner's needs in context, we leverage their PKG to provide the LLM with a model of the learner. To ensure a PKG-based Question Generation that generates questions that are relevant to the current learner's context, we carefully designed a prompt such that the LLM generates questions about the DNU concepts that are only based on the text provided from the current slide or the MCs contained in the slide.
When a learner marks a concept as DNU, the system triggers a \textit{graph-guided retrieval} function that retrieves information from the learner's PKG stored in the Neo4j database, including the DNU concept, the slide text (slide\_text) that contains the DNU concept, and other MCs that the slide contains (slide\_concepts), as shown in Figure \ref{pipeline-1-fig} (\textbf{A}). This knowledge is provided in a zero-shot Question Generation prompt template for a GPT 3.5-turbo LLM model (Figure \ref{pipeline-1-fig}, \textbf{P1}). To overcome challenges associated with Question Generation using LLMs, the prompt template is designed to follow some rules (qg\_rules), such as not to repeat questions that are semantically similar. To ensure that learners interact with the questions based on their importance to the slide, we developed \textit{graph-based re-ranking} of the questions by computing their embeddings using a sentence-transformer model and ranking them based on the similarity of their embeddings to the embedding of the slide text (Figure \ref{pipeline-1-fig}, \textbf{B}).
\subsection{EduKG-based Question Answering}
To answer a learner's selected question, we perform \textit{graph-guided retrieval} using the EduKG. To this end, the Wikipedia articles of the MCs in the current slide are chunked into a set of paragraphs and each Wikipedia Paragraph (WP) is indexed as a node in the graph vector store set up in Neo4j. The \textit{graph-based indexing} process takes place by calculating the vector embedding of each paragraph using a sentence-transformers language model.  
When the learner selects a question, the question is also transformed to a vector embedding and the most similar WPs to the questions are retrieved based on the cosine similarity, as shown in Figure \ref{pipeline-1-fig} (\textbf{C}).
After retrieving the most relevant Wikipedia contexts to the question (Figure \ref{pipeline-1-fig}, \textbf{D}), they are injected into the Extracive QA prompt template shown in Figure \ref{pipeline-1-fig} (\textbf{P2}). This prompt should cause the LLM's answer to be strictly based on extracted Wikipedia contexts, thus reducing hallucinations.
In case that the LLM does not generate answers (i.e., the MC WPs do not contain an answer to the selected question), we leverage the EduKG to get the RCs to the MCs in the slide. For example, the answer to a question such as 'What is parameter tuning in Machine Learning?' might not be in the WPs of the MC 'Machine Learning', however, it can be found in an RC of 'Machine Learning', such as 'Hyperparameter Optimization' because it is a more specialized topic that addresses this question. To achieve this, we load the full Wikipedia Articles (WA) of the RCs but without embedding. We avoid indexing the WAs of RCs in the graph vector store as each MC might have hundreds of RCs which would cause the \textit{graph-based indexing} to be slower. To retrieve answers from WAs of RCs, we employ an LLM retriever that is prompted using the EduKG retrieval prompt template (Figure \ref{pipeline-1-fig}, \textbf{P3}) to perform \textit{graph-guided retrieval} by traversing the EduKG and reasoning which RCs might contain an answer to the question (Figure \ref{pipeline-1-fig}, \textbf{E}). The WA of the retrieved RC is then provided back to the Extractive QA prompt template \textbf{P2} to extract answers from them.
After extracting the answers from the given contexts, the answers are provided to the learners along with citations of the resources (Figure \ref{pipeline-1-fig}, \textbf{F}). By clicking on the answer, the learner is redirected to the source with the answer highlighted as illustrated in Figure \ref{pipeline-1-fig} (\textbf{G}). This can ensure that learners explore beyond the answers provided to them by the LLM.
\section{Evaluation} \label{evaluation} In this section, we present the results of the human evaluation for the PKG-based Question Generation and EduKG-based Question Answering pipelines, in terms of linguistic and task-oriented dimensions. 
\subsection{PKG-based Question Generation Evaluation} We asked 3 instructors who deliver MOOCs through CourseMapper to evaluate the PKG-based Question Generation pipeline according to linguistic and task-oriented criteria. The evaluators are a Professor and two Teaching Assistants at the local university. The evaluation was carried out over three different MOOCs according to the instructor's area of expertise. The topics of the MOOCs were Learning Analytics (LA), Human-Computer Interaction (HCI) and Web Technologies (WT). Each instructor was asked to interact with the pipeline, in which they would select an LM from their MOOC. Then, they were asked to find slides that learners would find challenging. Then, they were asked to select any MC as DNU. For every DNU concept, they were asked to evaluate all recommended questions. The process was repeated for several DNU concepts in every MOOC. For LA and WT, 6 DNU concepts and 30 Question-Answer pairs were evaluated for each. For HCI, 8 DNU concepts and 40 Question-Answer pairs were evaluated.

We follow an evaluation framework similar to Fu et al. \cite{fu2024qgeval}. The framework provides seven dimensions for evaluating Question Generation models. The dimensions are divided into linguistic and task-oriented dimensions. Linguistic dimensions are Fluency (Flu.), Clarity (Clar.), and Conciseness (Conc.). The task-oriented dimensions are Relevance (Rel.), Consistency (Cons.), Answerability (Ans.), and Answer Consistency (AnsC.). Relevance is the most prevalent metric in evaluating Question Generation \cite{fu2024qgeval}. 
In the context of our pipeline, Relevance plays an important role since the main purpose of PKG-based Question Generation is to personalize the learning experience for every learner. This can be validated by measuring the relevance of the question to the chosen slide \((Rel_{slide}\)) and DNU concept (\(Rel_{dnuconcept}\)). The other task-oriented dimensions are all particularly defined to address challenges in Question Generation for reading comprehension tasks, which is not the focus of this work; therefore, we refrain from evaluating questions according to them. The five dimensions that we used, namely Flu., Clar., Conc., \(Rel_{slide}\), and \(Rel_{dnuconcept}\) are evaluated on a scale 1 to 3, the higher being better. 
Table \ref{tab:mooc_comparison} shows the results of the evaluation of the PKG-based Question Generation for the three different MOOCs. The evaluation shows very promising results with regards to both linguistic aspects and relevance. Our results show that the pipeline does not face linguistic challenges. In terms of relevance, the pipeline scores in general high, which is very promising as it shows that it can effectively support learners to ask the right questions in context, according to their needs. 
Furthermore, the instructors were very impressed with the level of detail in the questions and the level of relevance provided with regard to every slide and every DNU concept, and they showed an interest in deploying the feature into their MOOCs in the future. 
\renewcommand{\arraystretch}{1.7}
\begin{table}[h]
    \centering
    \caption{Comparison of PKG-based Question Generation on linguistic and relevance evaluation dimensions across different MOOCs}
    \begin{tabular}{|l|c|c|c|c|c|c|}
        \hline
        \textbf{MOOCs} & \textbf{Flu.} & \textbf{Clar.} & \textbf{Conc.} & \(\mathbf{Rel_{slide}}\) & \(\mathbf{Rel_{dnuconcept}}\) & \textbf{Avg.}\\
        \hline
        Learning Analytics  & 2.967 & 2.867 & 2.967 & 3.000 & 3.000 & 2.961  \\
        \hline
        Human-Computer Interaction  & 3.000 & 2.875 & 3.000 & 2.725 & 2.55 & 2.83 \\
        \hline
        Web Technologies  & 2.969 & 2.750 & 2.943 & 2.875 & 2.496 & 2.806 \\
        \hline
        \textbf{Weighted Avg.}  & 2.981 & 2.835 & 2.973 & 2.853& 2.668& 2.862 \\
        \hline
    \end{tabular}
    \label{tab:mooc_comparison}
\end{table}
\subsection{EduKG-based Question Answering Evaluation}
We asked the instructors to review the accuracy of the responses they receive on each question they select. While, there are other metrics to evaluate Question Answering, we choose accuracy in order to align with the evaluations of other RAG systems used in education \cite{liu2024hita,liu2024teaching}, which define accuracy as either correct or incorrect. Table \ref{tab:edukg_accuracy} presents the results of the perceived accuracy of EduKG-based Question Answering for the three different MOOCs. These results show the challenging aspect of EduKG-based Question Answering. According to the instructors, most answers were considered too abstract or not directly to the point which leads them to being identified as incorrect. By reviewing the data, we find several possible reasons for this. The EduKG is constructed of Wikipedia articles and the LLM is prompted to extract the most relevant information from the retrieved Wikipedia articles as it is with no further reasoning. While this might lead to higher levels of trust, as the retrieved text is exactly as in Wikipedia, it leads the LLM to generate responses that are abstract. Another reason could be the lack of disambiguation capabilities of the retrieval process. For example, in the MOOC HCI, one of the evaluated DNU concepts was 'Emergency Exit' in reference to a concept in User Experience (UX). However, the retriever was incapable to distinguish it from the 'Emergency Exit' for buildings.
In general, the instructors found the pipeline to be highly interactive. According to their comments, they believe that with further modifications to the EduKG-based Question Answering, the pipeline can be an essential tool for learners and instructors to have a more structured and personalized learning experience in MOOCs.
\begin{table}[h]
    \centering
    \caption{Accuracy of responses of the EduKG-based Question Answering in MOOCs}
    \renewcommand{\arraystretch}{1.7} 
    \resizebox{\textwidth}{!}{ 
    \large
    \begin{tabular}{|l|c|c|}
        \hline
        \textbf{MOOCs} & \textbf{Accuracy (\%)} & \textbf{Number of Correct Answers / Total Number of Answers} \\
        \hline
        Learning Analytics & 56.67 & 17/30 \\
        \hline
        Human-Computer Interaction & 45.00 & 18/40 \\
        \hline
        Web Technologies & 33.33 & 10/30 \\
        \hline
        \textbf{Weighted Average} & 45.00 & 45/100 \\
        \hline
    \end{tabular}
    } 
    \label{tab:edukg_accuracy}
\end{table}
\section{Conclusion and Future Work} \label{conclusion}
In this paper, we presented an innovative approach that leverages Educational Knowledge Graphs (EduKGs) and Personal Knowledge Graphs (PKGs) to implement a
Graph RAG pipeline that can guide learners in understanding new knowledge
concepts in the MOOC platform CourseMapper. The evaluation results show the potential of in-context Question Generation and Answering using Graph RAG in MOOCs. In particular, PKG-based Question Generation was perceived as effective to guide learners to learn new concepts in context. However, EduKG-based Question Answering still requires further enhancements to improve its accuracy and reliability. In this aspect, the first modification could be constructing the EduKG using further external learning resources that might hold more details to knowledge concepts than Wikipedia articles such as the learning material of other MOOCs, scientific literature or even recommended videos about the concepts. 
The second modification could be to review further methods that would allow the LLMs to use the provided resources as evidence but still give it room to perform reasoning and provide further explanations.
For example, the chain-of-thought method can allow the LLM to leverage an EduKG as a structured data source to perform advanced reasoning that is still relevant to the content of the MOOC and supported by evidence from the EduKG. 

\subsubsection{\discintname}
The authors have no competing interests to declare that are
relevant to the content of this article.
%
%
%
\bibliographystyle{splncs04}
\bibliography{sample_ceur}
%


\end{document}